\documentclass[10pt,twocolumn,letterpaper]{article}

\usepackage{iccv}
\usepackage{times}
\usepackage{epsfig}
\usepackage{graphicx}
\usepackage{amsmath}
\usepackage{amssymb}

\usepackage{multirow}
\usepackage{graphicx}
\usepackage{tabularx}
\usepackage{amsmath}
\usepackage{amssymb}
\usepackage{booktabs}
\usepackage{adjustbox}
\newcommand*\rot{\rotatebox{90}}
\usepackage{xcolor,colortbl}
\usepackage[accsupp]{axessibility}
\usepackage{enumitem}
\definecolor{amber}{rgb}{1.0, 0.75, 0.0}
\usepackage{algorithm}
\usepackage{listings}
\usepackage{bm}
\usepackage{tabularray}
\iccvfinalcopy 

\def\iccvPaperID{2172} 

\ificcvfinal\fi

\begin{document}

\title{Misalign, Contrast then Distill:\\Rethinking Misalignments in Language-Image Pretraining}

\author{
Bumsoo Kim\thanks{Correspondence to: \tt\footnotesize{bumsoo.kim@lgresearch.ai}}\hspace{0.4cm}
Yeonsik Jo\hspace{0.4cm}
Jinhyung Kim\hspace{0.4cm}
Seunghwan Kim\vspace{0.2cm}
\\LG AI Research
}

\maketitle
\ificcvfinal\thispagestyle{empty}\fi

\begin{abstract}
   Contrastive Language-Image Pretraining has emerged as a prominent approach for training vision and text encoders with uncurated image-text pairs from the web.
   To enhance data-efficiency, recent efforts have introduced additional supervision terms that involve random-augmented views of the image.
   However, since the image augmentation process is unaware of its text counterpart, this procedure could cause various degrees of image-text misalignments during training.
   Prior methods either disregarded this discrepancy or introduced external models to mitigate the impact of misalignments during training.
   In contrast, we propose a novel metric learning approach that capitalizes on these misalignments as an additional training source, which we term ``Misalign, Contrast then Distill (MCD)".
   Unlike previous methods that treat augmented images and their text counterparts as simple positive pairs, MCD predicts the continuous scales of misalignment caused by the augmentation.
   Our extensive experimental results show that our proposed MCD achieves state-of-the-art transferability in multiple classification and retrieval downstream datasets.
\end{abstract}

\section{Introduction}

Recent advances in deep learning have shown that image representations trained with large-scale uncurated natural language supervision shows powerful transferability to various downstream tasks~\cite{jia2021scaling,radford2021clip}.
A predominant paradigm in vision--language pre-training is to use a simple contrastive loss that makes the embedding of an image and its matching text description (positive pair) more similar to each other than other arbitrary image--text pairs (negative pairs)~\cite{oord2018representation}.
To achieve a more data-efficient training, following works actively capitalized on image random augmentation by: (i) joining language--image pretraining objectives with vision self-supervision terms~\cite{chen2020simclr,chen2021simsiam} between the augmented images~\cite{mu2022slip,li2022declip,lee2022uniclip} and (ii) involving more pairs of positive/negative supervisions between the augmented images and their original text description~\cite{li2022declip,lee2022uniclip}.

\begin{figure}
    \centering
    \includegraphics[width=\columnwidth]{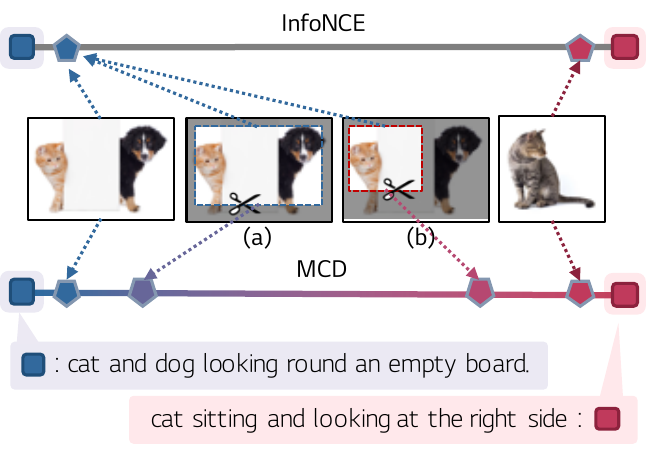}
    \caption{Conceptual illustration of contrastive language--image objectives of previous works (\ie, InfoNCE) and our MCD for (a) augmentation that doesn't harm the correspondence with its description and (b) augmentation that does.
    Previous works either disregard these misalignments~\cite{li2022declip} or leverage external models~\cite{lee2022uniclip,gao2022pyramidclip} to mitigate their impact.
    On the other hand, MCD uses the continuous degree of misalignments caused by random image augmentation as a useful source for training various levels of alignments between images and their text descriptions.
    }
    \label{fig:teaser}
    \vspace{-5pt}
\end{figure}

However, since the random image augmentation process is unaware of its corresponding text, it often results in the augmented image view to be \textit{misaligned} with its description (see (b) in Fig.\ref{fig:teaser}).
These misalignments behave as noisy training signals for the contrastive loss in language--image pretraining, thus causing performance degradation if not properly attended~\cite{mu2022slip}.
To mitigate this issue, recent works have used additional augmentation embeddings~\cite{lee2022uniclip} or heavy external off-the-shelf object detectors and summary extractors~\cite{gao2022pyramidclip} to match the alignment during training.
Though being straightforward and showing strong performance, these works are limited in that they add unnecessary burden in both training and inference.

Based on this observation, we start with a simple question: ``Instead of treating misalignments as noise to eliminate, can we rather harness them as a training source for language-image pretraining?".
To this end, we propose MCD (\ie, Misalign, Contrast then Distill), a novel training framework that leverages the various levels of misalignments between random augmented images and its text description during training.

\begin{figure}
    \centering
    \includegraphics[width=\columnwidth]{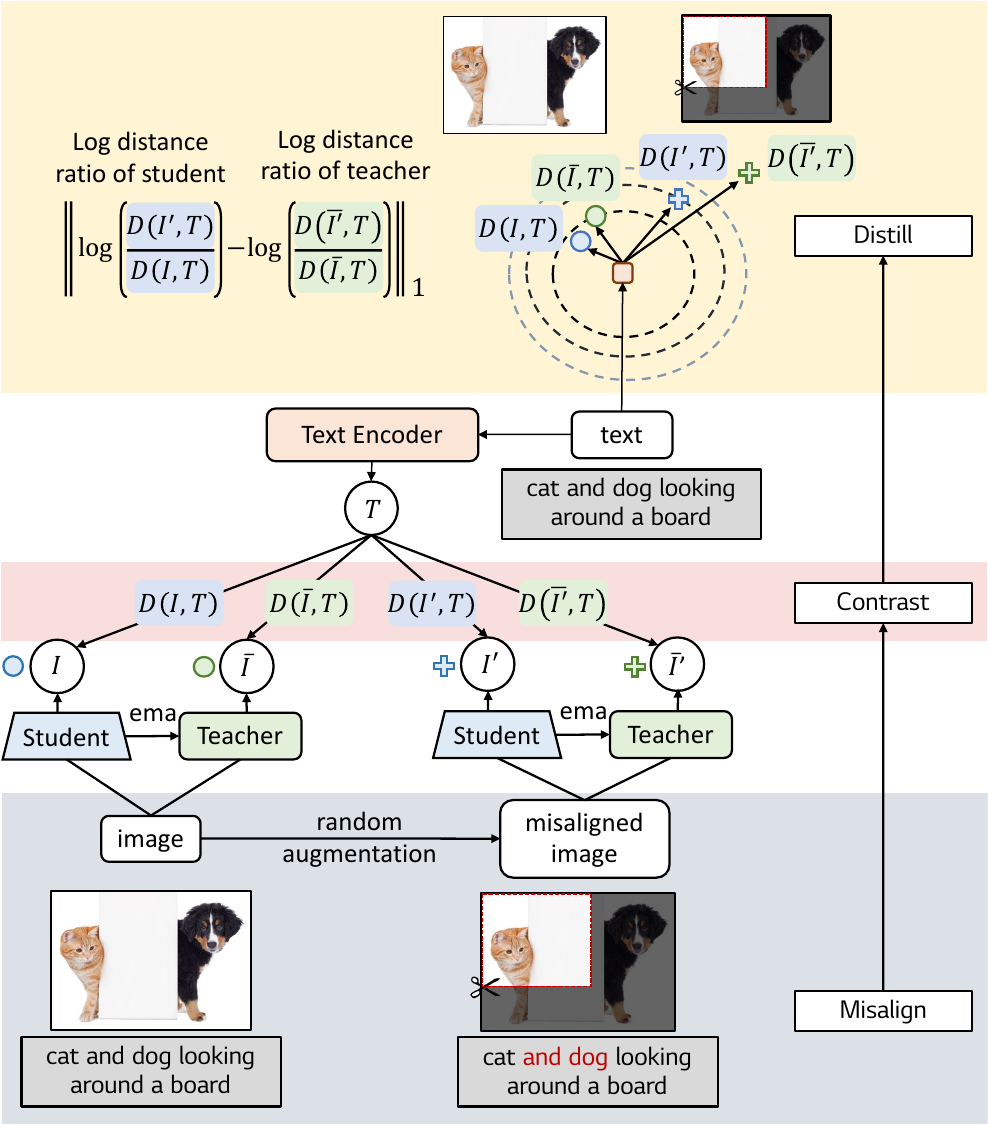}
    \caption{Overview of MCD. Our MCD consist of three steps: (i) text-agnostic random augmentation of the image that causes various levels of misalignment with the corresponding text, (ii) learning the distance between the image/augmented image and text with contrastive objectives then (iii) distill the log-ratio of the image-text distance between the original and the augmented image.
    }
    \label{fig:fig_overview}
\end{figure}

MCD consists of three steps (see Fig~\ref{fig:fig_overview} for an overview illustration of MCD):
First, we conduct random augmentation on the image that causes various levels of misalignments (or not at all) with its text counterpart (\textbf{Misalign}).
Then, we project all the participants (image, text, and augmented image) into an unified multimodal space, and learn the distance between all the image--text pairs with a contrastive objective (\textbf{Contrast}).
Finally, we use a teacher-student network where the student learns from the ``soft" distance between the text--original image (\ie, $D(\bar{I},T)$ in Fig~\ref{fig:fig_overview}) and text--augmented image (\ie, $D(\bar{I}',T)$ in Fig~\ref{fig:fig_overview}) of the momentum teacher with a log-ratio loss (\textbf{Distill}).
continuous scale of misalignment to the student model, enabling the student to learn from the various levels of misalignment that occur from the random augmentation during training time.


Our contribution of this paper is threefold:
\begin{itemize}
    \item We propose MCD, a novel training framework where we learn the continuous level of misalignment as a source for contrastive language-image pretraining.
    \item Our MCD outperforms state-of-the-art models across various single/multi-modal downstream datasets without adding additional parameters for inference or using external models to force the alignment.
    \item We propose three distillation strategies leveraging misalignments: i) misalignment between positive pairs, ii) misalignment between negative pairs, and iii) misalignment between noisy pairs. Extensive experiments show that all three strategies positively contributes to our final performance.
\end{itemize}

\section{Related Work}
Here, we present brief review of multi-modal representation learning, especially vision-language pre-training.

\subsection{Vision-language Pre-training}
Vision-Language Pre-training (VLP) trains a multi-modal model to learn joint representation of visual and textual information that can be transfer to various vision--language downstream tasks.
The success of VLP primarily relies on large-scale datasets which contains images and their corresponding descriptions, enabling the model to understand the semantic relationship between the pairs.
VLP includes two different group of models: 1) single-stream models~\cite{li2019visualbert,chen2020uniter,huang2019unicoder,li2020oscar,li2020unimo,lu2019vilbert,tan2019lxmert} which process an image and its associated text information in a shared backbone network and 2) dual-stream models~\cite{jia2021scaling,radford2021clip} which has two independent backbone networks for processing each modality.
In this work, we focuses on Contrastive Language Image Pretraining (CLIP~\cite{radford2021clip}), which is a type of dual-stream models trained with image-text contrastive loss where the image and its matching text description in the dataset as a positive pair and other unrelated pairs in a batch as negative pairs.

\subsection{Misalignments between Image-Text Pairs}
There are two different sources that cause misalignments in image-text pairs for VLP: misalignment that naturally occurs in image-text paired datasets, and misalignments caused by random image augmentation.

\paragraph{Misalignment in Image-Text Pairs.}
Large scale image-text paired datasets for VLP are usually collected from the web thus can contain uncurated and noisy pairs which have weak relations.
This inevitably incurs misalignment between positive image-text pairs in the dataset misleading na\"ive contrastive language image objective.
Previous studies~\cite{li2021albef,Lu2022COTS,Andonian2022robust} have attempted to address the problem by knowledge distillation~\cite{hinton2015distilling} of soft image-text alignment matrix from momentum teacher network to the student network via KL divergence loss.
Our proposed method stands apart from previous approaches due to its element-wise log-ratio loss for distillation. Element-wise loss lessens the dependence on training hyperparameters like batch size and temperature for KL loss. Furthermore, it enables the model to harness the various levels of individual misalignments of each sample from random augmentation or label noise.

\paragraph{Misalignment by Augmentations.}
View-based self-supervised learning, in which models
are trained to represent views or augmentations of the same image similarly, has yielded strong results across a variety of different formulations.
Consecutive work of CLIP (i.e., SLIP~\cite{mu2022slip}) initially introduced supervision between random augmented image views (\eg, cropping, gray-scale, jittering, gaussian blur, horizontal flipping, etc.).
Since only image-image supervision was given in a separately embedded space, augmentation was not a factor for misalignment.
To include more positive/negative pairs for the image-text contrastive objective, following works~\cite{li2022declip,lee2022uniclip} also introduced InfoNCE loss between the augmented views and the text pairs.
However, since the text description is unaware of the random augmentations, there is a high chance that misalignments occur during training.
SLIP shows that na\"ive application of contrastive loss to these misaligned pairs results in suboptimal performance.
Previous work have either ignored this misalignment~\cite{li2022declip} or addressed this issue with an additional encoder that encodes the one-hot information of which augmentation has been applied during training~\cite{lee2022uniclip}.
PyramidCLIP~\cite{gao2022pyramidclip} addressed this issue with external off-the-shelf object detectors and summary extractors.
HiCLIP~\cite{geng2023hiclip} captured the hierarchical nature of high-level on unsupervised manner with tree Transformer~\cite{wang2019treeTransformer}.
MCD incorporate the misalignment information as a source of training via novel log-ratio loss without introducing any external module.
\section{Preliminary}

In our preliminary, we revisit the basic form of Contrastive Language-Image Pretraining (\ie, CLIP~\cite{radford2021clip}).
CLIP features a dual-encoder architecture where the image encoder $f_I$ and text encoder $f_T$ are jointly trained with a contrastive objective $\mathcal{L}_{\text{CLIP}}$.

\paragraph{InfoNCE Loss}
Given $N$ image-text pairs $\{(x_i^I,x_i^T)\}_{i=1}^N$, we define a similarity matrix $S$ whose $i$-th row and the $j$-th column is the cosine similarity between the projected representations of the $i$-th text $T_i$ and the $j$-th image $I_j$ (\ie, $T_i=f_T(x^T_i)$, $I_j=f_I(x^I_j)$), written as:
\begin{equation}
\label{eq:align}
    S_{ij}=\mbox{sim}(T_i, I_j),
\end{equation}
where sim$(\cdot, \cdot)$ is cosine similarity.
In CLIP~\cite{radford2021clip}, the encoded image features $I$ and text features $T$ are projected to the same dimension where the embeddings for matching image-text pairs are pulled together while embeddings for arbitrary pairs are pushed apart with the InfoNCE loss~\cite{oord2018representation}.
Given the similarity matrix $S$, the InfoNCE loss $\mathcal{L}_N$ is rewritten as:
\begin{equation}
\label{eq:infoNCE}
    \mathcal{L}_{N}(S)=-\frac{1}{N}\sum_{i=1}^N\log{\frac{\exp{\big(S_{ii}/\tau\big)}}{\sum_{j=1}^N\exp{\big(S_{ij}/\tau\big)}}},
\end{equation}
where $\tau$ is a learnable temperature variable.
The overall loss of clip $\mathcal{L}_{\text{CLIP}}$ is written as:
\begin{equation}
\label{eq:CLIP}
    \mathcal{L}_{\text{CLIP}}=\frac{1}{2}\bigg(\mathcal{L}_{N}(S)+\mathcal{L}_{N}(S^T)\bigg).
\end{equation}

\paragraph{Augmentations and Misalignment.}
To include more positive/negative pairs for the contrastive objective, following works~\cite{li2022declip,lee2022uniclip} introduced InfoNCE loss between the random augmented image views and the text description for the original image.
Let $\mathcal{A}$ a function for random image augmentation that includes randomly applying cropping, gray-scale, jittering, gaussian blur, horizontal flipping following SimCLR~\cite{chen2020simclr}. 
$I_{j'}$ is the encoded image feature of augmented view of $j$-th image(\eg $I_{j'} = f_I(\mathcal{A}(x_j^I))$).
Let $S_{ij}'=\text{sim}(T_i,I_{j'})$ represent the similarity between the augmented view of $j$-th image and $i$-th text. We denote the matrix of these similarity as $S'$ for concise notation.
Then, InfoNCE loss between the augmented image and the text is
\begin{equation}
\mathcal{L}_\text{CLIP}'=\frac{1}{2}\big(\mathcal{L}_{N}(S')+\mathcal{L}_{N}(S'^T)\big).
\end{equation}

As the random augmentation function $\mathcal{A}$ is independent to the corresponding text description, the augmented view $I_{j'}$ is likely to exhibit misalignment with text $T_i$.
This hypothesis is consistent with the finding of SLIP~\cite{mu2022slip}, which demonstrated that introducing augmentation (particularly resize crop, and flip) to CLIP actually resulted in a performance decrease.
Previous works~\cite{mu2022slip, li2022declip} have sidestepped the utilization of augmented view in CLIP by substituting infoNCE with self-supervised learning loss(\eg SimCLR~\cite{chen2020simclr}) between images. These approaches have limitation in fully capturing the essence of multi-modal learning.
\section{Method}

In this section, we introduce MCD (Misalign, Contrast then Distill), a novel training framework for language--image pretraining using misalignments as \textit{continuous} labels for learning the distance between image--text pairs.
\subsection{Misalign}
\begin{figure}
    \centering
    \includegraphics[width=\columnwidth]{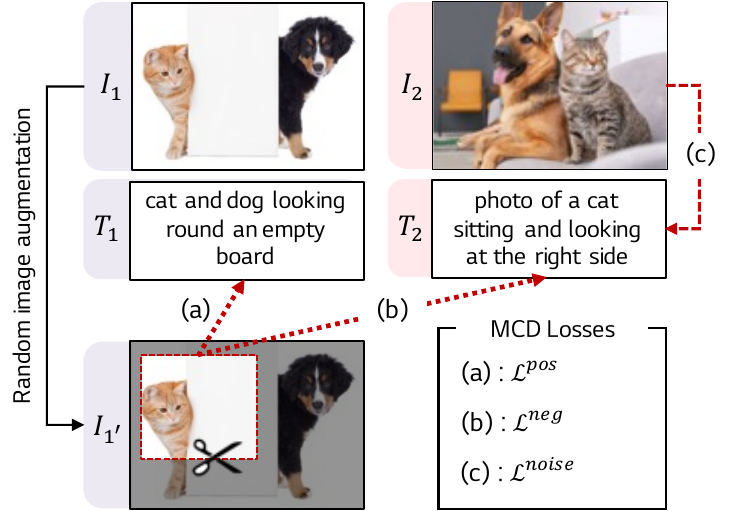}
    \vspace{5pt}
    \caption{MCD loss for various case of misalignments that occur when applying random image augmentation in language--image pretraining with. In this paper, we elaborate three scenarios: (a) misalignment caused by augmenting the image of an original positive pair, which is addressed by $\mathcal{L}^\text{pos}$ (b) augmentation that mistakenly cause positive alignment between negative pairs (c) misalignment that already exists within the dataset.
    }
    \label{fig:fig_overall}
\end{figure}

The first step of MCD is to apply text-agnostic random image augmentations (\eg, random crop, random flip, grayscale, etc.) to create various levels of misalignments between the images and its description.
Here, we elaborate on three distinct scenarios of misalignment that can arise during the augmentation process, hindering the contrastive loss to learn proper distance between image--text pairs: (i) Text-agnostic random augmentation can cause misalignments in positive image-text pairs.  (ii) The random-augmentation can mistakenly cause positive signals between negative pairs. (iii) Misalignments can already exist innately within the original image-text pair. A detailed illustration of each cases are provided in Fig~\ref{fig:fig_overall}.

\subsection{Contrast}
In the second step, MCD initially learns the distance metric between image--text pairs with a contrastive objective.
Motivated by~\cite{lee2022uniclip}, we project both image and text modalities to a unified space and use all the positive pairs and negative pairs of both modalities.
For an $i$-th embedding $z_i$ in a batch of embeddings $\{z_i\}_{i=1}^{3N}$ that includes $N$ image samples, $N$ text samples, and $N$ random augmented image samples, let $\mathcal{P}_i$ and $\mathcal{N}_i$ each denote the set of all positive sample indices of the $i$-th sample including $i$ itself and the set of all negative sample indices of the $i$-th sample.
Then, the contrastive loss for the multiple positives and multiple negatives for sample $i$ can be written as
\begin{equation}
\label{eq:MP-NCE}
\small
\mathcal{L}^C_{i}= \mathbb{E}_{p \in \mathcal{P}_i} \left[-\log \frac{\text{sim}(z_i, z_p)}{\text{sim}(z_i,z_p)+\sum_{n \in \mathcal{N}_i}\text{sim}(z_i,z_n)} \right].
\end{equation}
However, without encoding augmentation information the image-text contrastive loss is prone to the three aforementioned issues.
We address these three issues with a teacher-student model where the continuous distance between the image--text and augmented image--text of the teacher model is distilled to the student model.

\begin{algorithm}[!htbp]
\caption{MCD Pseudocode}
\label{alg:code}
\definecolor{codeblue}{rgb}{0.25,0.5,0.5}
\definecolor{codekw}{rgb}{0.85, 0.18, 0.50}
\begin{lstlisting}[language=python,escapeinside=;;]
# fm, fs: image encoders (teacher, student)
# ft: text encoder
# fa: random-augmentation function
# N : batch size

def D(sim): # cosine sim -> L2 distance
    return 2 - 2 * sim + 1e-6
    
def MCD(img, txt):
    # image, text encodings
    aug = fa(img)

    # normalized projection embeddings
    zm, zs, zt = fm(img), fs(img), ft(txt)
    zam, zas = fm(aug), fs(aug) # misalign

    # distance (contrast)
    di_t, di_s = D(zm @ zt.T), D(zs @ zt.T)
    da_t, da_s = D(zam @ zt.T), D(zas @ zt.T)

    # distill
    pos_l, neg_l, noise_l = 0, 0, 0
    for i in range(N):
        lr_s1 = log(da_s[i,i]/di_s[i,i])
        lr_t1 = log(da_t[i,i]/di_t[i,i])

        # positive pairs
        pos_l += abs(lr_s1 - lr_t1) / N

        for j in range(N):
            if i==j: continue
            lr_s2 = log(da_s[i,j]/di_s[i,i])
            lr_t2 = log(da_t[i,j]/di_t[i,i])

            # negative pairs
            neg_l += abs(lr_s2 - lr_t2)/(N*(N-1))

            lr_s3 = log(di_s[j,j]/di_s[i,i])
            lr_t3 = log(di_t[j,j]/di_t[i,i])

            # noisy pairs
            noise_l += abs(lr_s3 - lr_t3)/(N*(N-1))

    return pos_l + neg_l + noise_l
\end{lstlisting}
\end{algorithm}
\subsection{Distill}
Knowledge distillation, introduced by Hinton et al.~\cite{hinton2015distilling}, is a learning paradigm where we train the student network to mimic the ``soft" labels predicted from the teacher network.
In MCD, we train the student network with the continuous image--text distance predicted by the teacher network.
Psuedocode for the distillation in MCD is provided in Algorithm \ref{alg:code}.

\paragraph{Log-Ratio Loss for Image--Text Distance.}
Given a student $f_I$ and a momentum teacher $\bar{f}_I$, we propose to use log-ratio loss~\cite{kim2019logratio} on image-text similarities that aims to approximate the ratio of similarity distances by the ratio of image-text misalignments in the learned embedding space.
Specifically, we define the loss function as
\begin{equation}
\ell_{lr}(i, j; \alpha, \beta) =\biggl\Vert\log{\frac{D\bigl(I_\alpha, T_\beta\bigl)}{D\bigl(I_i, T_j\bigl)}}-\log{
\frac{D\bigl(\bar{I}_\alpha, T_\beta\bigl)}{D\bigl(\bar{I}_i, T_j\bigl)}}\biggl\Vert_1,
\label{eq:base_logratio}
\end{equation}
where $\Vert \cdot \Vert_1$ is $\ell_1$ loss, $D(\cdot, \cdot)$ is distance function.
We utilize Euclidean distance between the projected representations as distance function. 
Since the embedded vectors comprising the image $I_i, \bar{I}_i$ and text $T_j$ are L2-normalized, Euclidean distance operates as a proxy for cosine similarity $D(I_i, T_j) = 2(1-S_{ij})$. 
This log-ratio loss approximates the degree of misalignment, which is measured as the ratio of two image-text pairs.
By leveraging this measure, we aim to promote a coherent and continuous embedding space. Accordingly, our student encoder is trained under the guidance of the momentum teacher with the incorporation of this degree of misalignment.
By establishing these pairs of log-ratio, we enable the handling of diverse forms of misalignment. We present three distinct index setups that correspond to different types of misalignment.

\paragraph{Misalignment in Positive pairs.}
First, we define distance pair for misalignment between the original image-text pair. Let $i'$ denote the index of augmented image sample. On Eq.(\ref{eq:MP-NCE}), $i'$-th image sample and $i$-th text sample serve as positive pair. However, random augmentation can occasionally transform positive pair into negative pair. To account for such transformations, we intend to utilize the log-ratio between original pairs and augmented pairs. This allows us to capture these shifts and incorporate them into the learning process effectively.

\begin{equation}
\label{eq:lr_postive}
    \mathcal{L}^{\text{pos}} = \mathbb{E}_{i=1,...,N} \left[\ell_{lr}(i, i; i', i)\right],
\end{equation}
where $i'$ is the index of augmented $i$-th sample.

\paragraph{Misalignment in Negative pairs.}
Augmented images can possess relevance with different text, which would normally be considered negative pairs in Eq.(\ref{eq:MP-NCE}).
However, the log-ratio obtained by our momentum teacher alleviates our model from mistakenly pushing the embedding of relevant texts away.
\begin{equation}    
\mathcal{L}^{\text{neg}} = \mathbb{E}_{\substack{i,j=1,2,...,N\\i\ne j}}\left[ \ell_{lr}(i,i;j',i) \right],
\end{equation}
where $j'$ is the index of augmented $j$-th sample.

\paragraph{Misalignment in Noisy pairs.}
Original image-text pairs obtained from the web may contain either noisy images or descriptions.
While these pairs are normally treated as positive pairs under contrastive loss, we propose a loss for noisy pairs where the noisy labels are trained to have larger distance than matching image--text pairs.
\begin{equation}
\mathcal{L}^{\text{noisy}} = \mathbb{E}_{\substack{i,j=1,2,...,N\\i\ne j}}\left[ \ell_{lr}(i,i;j,j) \right].
\end{equation}

\paragraph{Distillation Loss.}
The full training objective of MCD distillation $\mathcal{L}^D$ is the sum of the three distillation terms, which is written as
\begin{equation}
\mathcal{L}^\text{D}=\mathcal{L}^\text{pos}+\mathcal{L}^\text{neg}+\mathcal{L}^\text{noisy}.
\end{equation}
\subsection{Training MCD}
In this section, we explain the details of MCD training.
The training objective for the text encoder and student image encoder for MCD consists of the three objectives: contrastive loss $\mathcal{L}^C$ in Eq~(\ref{eq:MP-NCE}) for initial image--text distance learning, distillation loss $\mathcal{L}^D$ for the three misalignment scenarios, and loss for masked language modeling $\mathcal{L}^{\text{MLM}}$.
The parameters for the teacher image encoder are momentum updated.

\paragraph{MLM Loss.}
Following previous work in literature~\cite{li2022declip,lee2022uniclip,li2021albef}, we randomly mask out the input tokens with a probability of 15\% and replace them with the
special token \texttt{[MASK]}\footnote{Following BERT, the replacement is done with either the \texttt{[MASK]} token (80\%), another random token within the dictionary (10\%), or left unchanged (10\%).}.
Let $p
^{\text{msk}}$ and $y^{\text{msk}}$ each denote the set of model’s predicted
probability for the masked tokens and the set of ground-truth vocabulary index for the tokens, respectively.
Then, MLM loss is written as:
\begin{equation}
\label{eq:MLM}
\mathcal{L}^\text{MLM}=\mathbb{E}_{p \in p^{\text{msk}}, y \in y^{\text{msk}}} \left[\mbox{CE}(p, y)\right],
\end{equation}
where CE denotes Cross Entropy loss.

\paragraph{Momentum Teacher Update.}
Let $\theta_{f_I}$, $\theta_{\bar{f}_I}$ be the parameter of the student encoder and momentum teacher, respectively.
For the $t$-th step, we update $\theta^{(t)}_{\bar{f}_I}$ of the momentum teacher according to the following:
\begin{equation}
    \theta_{\bar{f}_I}^{(t)} = m \theta_{\bar{f}_I}^{(t-1)} + (1-m) \theta_{f_I}^{(t)},
\end{equation}
where $m$ denotes the momentum parameter. We use $m=0.994$ in our experiments, where $m$ grows in a cosine schedule to $1$ at the end of training.

\paragraph{Progressive Distillation.}
As the training proceeds, InfoNCE loss conflicts with our misalignment loss.
InfoNCE between pair $I_j'$ and $T_i$ forces the embedding to pull regardless of its degree of misalignment.
In early stages of training, the model needs to learn how to discriminate positive or negative pairs with a hard label.
However, as the training progresses, the log-ratio loss delicately models the distance between the various misalignments occurred by augmentations or innately existing in the original image-text pair.
Therefore, we progressively diminish the contribution of InfoNCE loss involving augmented views.

\paragraph{MCD Loss.}
The final loss for MCD is written as:
\begin{equation}
\mathcal{L}^\text{MCD}=\mathcal{L}^\text{C}+\alpha\cdot\mathcal{L}^\text{D}+\beta\cdot\mathcal{L}^\text{MLM},
\end{equation}
where $\alpha=0$ progressively increases on a cosine schedule to 1, and $\beta=0.2$.

\subsection{MCD Inference}
MCD is based on a teacher-student network, thus the student $f_I$ and momentum teacher $\bar{f_I}$ is obtained after training.
Unlike previous work in literature that leverage the teacher network for inference~\cite{Tarvainen2017meanteacher,li2021albef}, we use the student network that is trained with both the contrastive loss and the log-ratio loss for image--text distance learning.
\section{Experiment}
In this section, we provide implementation details and experimental results with our MCD pretrained on two widely used image-text benchmark datasets (\eg, CC3M, YFCC15M) to validate the effectiveness of our proposed MCD on multiple downstream datasets including classification and image--text retrieval.

\subsection{Implementation Details and Datasets}
For implementation details, our work is built on top of the open-source SLIP codebase~\cite{mu2022slip}\footnote{https://github.com/facebookresearch/SLIP \label{slip_github}}.
For DECLIP~\cite{li2022declip}, we follow the implementation details of the official code release\footnote{https://github.com/Sense-GVT/DeCLIP}.
The performance on GPU-machine runs for CLIP and SLIP follows the exact implementation details upon this codebase.
Since MCD features both the momentum teacher image encoder $\bar{f_I}$ and student $f_I$, we conduct the following experiment section with $\bar{f_I}$ based on empirical results.
All of our models are pretrained in 16$\times$ A100 GPUs.
For CC3M, All models are trained with a ViT-B/16 backbone with a learning rate of 5e-4 and weight decay of 0.5.
For YFCC15M, we train the model with ViT-B/32 backbone, batch size 4096, learning rate 1e-3, and weight decay 0.2.

\paragraph{Pretraining Datasets.}
To validate the effectiveness of MCD, we pretrain MCD on large-scale open-source datasets: YFCC (Yahoo Flickr Creative Commons) 15M~\cite{Thomee2016YFCC100M} and CC (Conceptual Captions) 3M~\cite{sharma2018cc3m}.

\paragraph{Downstream Datasets.}
Following CLIP~\cite{radford2021clip}, we evaluate the transferability of pretrained MCD on 11 widely used downstream datasets for classification (\ie, Oxford Pets~\cite{parkhi2012cats}, CIFAR-10, CIFAR-100~\cite{krizhevsky2009learning}, SUN397~\cite{xiao2016sun}, Food-101~\cite{bossard2014food}, Flowers~\cite{nilsback2008automated}, Cars~\cite{krause20133d}, Caltech-101~\cite{fei2004learning}, Aircraft~\cite{maji2013fine}, DTD~\cite{cimpoi2014describing}, ImageNet-1k~\cite{Russakovsky2014imagenet1k}).
We also transfer to image--text retrieval tasks on Flickr30K~\cite{plummer2015flickr30k} and MS-COCO Captions~\cite{chen2015microsoft} datasets.
The evaluation settings for each dataset are consistent with CLIP as in the open-source implementation\footref{slip_github}.
\subsection{MCD Pretraining on YFCC15M Dataset}

First, we pretrain MCD on YFCC15M and evaluate its transferability in single-modal (\eg, classification) and multi-modal (\eg, image--text retrieval) downstream tasks.
We compare the result against other state-of-the-art Contrastive Language-Image Pretraining approaches~\cite{radford2021clip, mu2022slip,li2022declip,lee2022uniclip} that utilizes various levels of supervision including vision self-supervision~\cite{chen2020simclr,chen2021simsiam}, text self-supervision~\cite{li2022declip}, memory queue~\cite{li2022declip}, and augmentation encoding~\cite{lee2022uniclip}.
All models are pretrained with a learning rate 1e-3 for 32 epochs unless mentioned otherwise.

\paragraph{Zero-shot Classification.}
\begin{table*}[!htbp] 
\centering
\small
    \begin{tabularx}{\textwidth}{lc c*{11} c}
        \toprule
        Method & \shortstack{Vision\\Encoder} & \rot{Oxford Pets} & \rot{CIFAR-10} & \rot{CIFAR-100} & \rot{SUN397} & \rot{Food-101} & \rot{Flowers} & \rot{Cars} & \rot{Caltech-101} & \rot{Aircraft} & \rot{DTD} & \rot{ImageNet} & \textbf{Average} \\\midrule
        \rowcolor[gray]{0.95}\multicolumn{14}{l}{\textit{Zero-shot Classification:}} \\ \\[-9pt]
        CLIP~\cite{radford2021clip} & \multirow{5}{*}{\hspace{13pt}ViT-B/32\hspace{13pt}} & 19.4 & 62.3 & 33.6 & 40.2 & 33.7 & 6.3 & 2.1 & 55.4 & 1.4 & 16.9 & 31.3 & \hspace{15pt}27.5\hspace{13pt} \\
        SLIP~\cite{mu2022slip} & & 28.3 & 72.2 & 45.3 & 45.1 & 44.7 & 6.8 & 2.9 & 65.9 & 1.9 & 21.8 & 38.3 & 33.9 \\
        DeCLIP~\cite{li2022declip} & & 30.2 & 72.1 & 39.7 & 51.6 & 46.9 & 7.1 & 3.9 & 70.1 & 2.5 & 24.2 & 41.2 & 35.4 \\
        UniCLIP~\cite{lee2022uniclip} & & 32.5 & 78.6 & 47.2 & 50.4 & 48.7 & \textbf{8.1} & 3.4 & 73.0 & 2.8 & 23.3 & 42.8 & 37.3 \\
        MCD (Ours) & & \cellcolor{blue!10}{\textbf{40.0}} & \cellcolor{blue!10}{\textbf{80.3}} & \cellcolor{blue!10}{\textbf{49.6}} & \cellcolor{blue!10}{\textbf{55.3}} & \cellcolor{blue!10}{\textbf{54.0}} & \cellcolor{blue!10}{7.9} & \cellcolor{blue!10}{\textbf{4.5}} & \cellcolor{blue!10}{\textbf{73.2}} & \cellcolor{blue!10}{\textbf{3.0}} & \cellcolor{blue!10}{\textbf{30.5}} & \cellcolor{blue!10}{\textbf{44.7}} & \cellcolor{blue!10}{\textbf{40.2}}\\
        \midrule
        \rowcolor[gray]{0.95}\multicolumn{14}{l}{\textit{Linear Probing:}} \\ \\[-9pt]
        CLIP~\cite{radford2021clip} & \multirow{5}{*}{ViT-B/32} & 71.2 & 89.2 & 72.1 & 70.1 & 71.4 & 93.2 & 34.9 & 84.3 & 29.7 & 60.9 & 61.1 & 67.1 \\
        SLIP~\cite{mu2022slip} & & 75.4 & 90.5 & 75.3 & 73.5 & 77.1 & 96.1 & 43.0 & 87.2 & 34.1 & 71.1 & 68.1 & 71.9 \\
        DeCLIP~\cite{li2022declip} & & 76.5 & 88.6 & 71.6 & 75.9 & 79.3 & 96.7 & 42.6 & 88.0 & 32.6 & 69.1& 69.2 & 71.8 \\
        UniCLIP~\cite{lee2022uniclip} & & 83.1 & 92.5 & 78.2 & 77.0 & 81.3 & 97.1 & \textbf{49.8} & 88.9 & 36.2 & 72.8 & 70.8 & 75.2 \\
        MCD (Ours) & & \cellcolor{blue!10}\textbf{85.6} & \cellcolor{blue!10}{\textbf{92.7}} & \cellcolor{blue!10}{\textbf{79.3}} & \cellcolor{blue!10}{\textbf{77.6}} & 
        \cellcolor{blue!10}{\textbf{81.7}} & 
        \cellcolor{blue!10}{\textbf{97.1}} & 
        \cellcolor{blue!10}{46.9} & 
        \cellcolor{blue!10}{\textbf{89.5}} & 
        \cellcolor{blue!10}{\textbf{36.6}} & 
        \cellcolor{blue!10}{\textbf{74.1}} & \cellcolor{blue!10}{\textbf{71.3}} & 
        \cellcolor{blue!10}{\textbf{75.7}} \\
        \bottomrule
    \end{tabularx}
    \vspace{1pt}
    \caption{Zero-shot image classification/linear probing performance on 11 downstream datasets with YFCC15M pretrained models. Note that DeCLIP~\cite{li2022declip} utilizes an external momentum queue while UniCLIP~\cite{lee2022uniclip} features the augmentation encoder during training.}
    \label{result:yfcc15m_cls}
\end{table*}
We evaluate the zero-shot classification performance on 11 downstream datasets for single-modal experiments.
Tab.~\ref{result:yfcc15m_cls} shows both the zero-shot classification and linear probing accuracy of CLIP variants~\cite{radford2021clip,mu2022slip,li2022declip,lee2022uniclip} pretrained on YFCC15M dataset and transferred to downstream classification datasets.
In test time, the learned text encoder $f_T$ synthesizes a zero-shot linear classifier by embedding the arbitrary categories of the test dataset.
As classes are in the form of a single word, we use prompts including the label (e.g., ``\texttt{a photo of a \{label\}}") as following CLIP~\cite{radford2021clip}.
Our MCD outperforms across a majority of the 11 datasets with a noticeable margin.
Note that even without additional augmented-aware network leveraged in UniCLIP~\cite{lee2022uniclip} or additional supervision terms such as text augmentation~\cite{wei2019eda}, masked language modeling~\cite{li2022declip} and memory queue, our MCD achieves state-of-the-art performance.

\paragraph{Linear Probing.}
To implement linear probe evaluation, we follow CLIP~\cite{radford2021clip} to train a logistic regression classifier on the frozen visual features extracted by the image encoder $f_I$.
Specifically, we train the logistic regression classifier using L-BFGS algorithm provided by scikit-learn with maximum 1,000 iterations, and report the corresponding metric for each dataset\footnote{https://github.com/facebookresearch/SLIP/blob/main/main\_linear.py}.
Parameters for L2 regularization are determined using hyperparameter sweep on the validation sets.
Standard cropping and flipping augmentations~\cite{szegedy2015going} are used for linear probing.
The bottom section of Tab. \ref{result:yfcc15m_cls} reports linear classification performances on the 11 downstream datasets.
Our proposed approach, MCD, has consistently outperformed previous baseline methods in zero-shot classification across multiple datasets, with only one exception.

\paragraph{Image--Text Retrieval.}
\begin{table*}[t]
    \centering
    \begin{tabularx}{\textwidth}{l ccc|ccc|ccc|ccc}
        \toprule
        & \multicolumn{6}{c}{Image-to-text retrieval} & \multicolumn{6}{c}{Text-to-image retrieval} \\
         & \multicolumn{3}{c}{Flickr30k} & \multicolumn{3}{c}{COCO Captions} & \multicolumn{3}{c}{Flickr30k} & \multicolumn{3}{c}{COCO Captions} \\
        Method\hspace{38pt}  & R@1 & R@5 & R@10 & R@1 & R@5 & R@10 & R@1 & R@5 & R@10 & R@1 & R@5 & R@10 \\
        \midrule
        \rowcolor[gray]{0.95}\multicolumn{13}{l}{\textit{Zero-shot retrieval:}} \\ \\[-9pt]
        CLIP~\cite{radford2021clip} & 34.9	& 63.9 & 75.9 & 20.8 & 43.9 & 55.7 & 23.4 & 47.2 & 58.9 & 13.0 & 31.7 & 42.7 \\
        SLIP~\cite{mu2022slip} & 47.8	& 76.5 & 85.9 & 27.7 & 52.6 & 63.9 & 32.3 & 58.7 & 68.8 & 18.2 & 39.2 & 51.0\\
        DeCLIP~\cite{li2022declip} & 51.4 & 80.2 & 88.9 & 28.3 & 53.2 & 64.5 & 34.3 & 60.3 & 70.7 & 18.4 & 39.6 & 51.4 \\
        UniCLIP~\cite{lee2022uniclip} & 52.3 & 81.6 & 89.0 & 32.0 & 57.7 & 69.2 & 34.8 & 62.0 & 72.0 & 20.2 & 43.2 & 54.4 \\
        MCD (Ours) & \cellcolor{blue!10}{\textbf{57.6}} & \cellcolor{blue!10}{\textbf{82.6}} & \cellcolor{blue!10}{\textbf{91.1}} & \cellcolor{blue!10}{\textbf{32.3}} & \cellcolor{blue!10}{\textbf{58.7}} & \cellcolor{blue!10}{\textbf{71.2}} & \cellcolor{blue!10}{\textbf{36.4}} & \cellcolor{blue!10}{\textbf{64.8}} & \cellcolor{blue!10}{\textbf{74.1}} & \cellcolor{blue!10}{\textbf{20.7}} & \cellcolor{blue!10}{\textbf{43.5}} & \cellcolor{blue!10}{\textbf{55.3}} \\
        \midrule
        \rowcolor[gray]{0.95}\multicolumn{13}{l}{\textit{Fine-tuned retrieval}} \\ \\[-9pt]
        CLIP~\cite{radford2021clip}  &  58.3 & 84.8 & 91.5 & 36.1 & 65.0 & 76.4 & 43.1 & 71.1 & 80.3 & 24.9 & 51.7 & 64.1 \\
        SLIP~\cite{mu2022slip} & 69.6 & 90.4 & 95.7 & 45.0 & 74.0 & 83.0 & 52.1 & 79.4 & 86.9 & 31.6 & 59.5 & 71.3 \\
        DeCLIP~\cite{li2022declip} & 75.6 & 93.0 & 96.6 & 48.7 & 77.3 & 86.2 & 57.8 & 83.3 & 90.3 & 34.2 & 63.1 & 74.6\\
        UniCLIP~\cite{lee2022uniclip} & 78.1 & 94.9 & 97.7 & 54.5 & 80.9 & 89.1 & 61.0 & 86.0 & 91.9 & 38.0 & 67.2 & 78.0 \\
        MCD (Ours) & \cellcolor{blue!10}{\textbf{79.3}} & \cellcolor{blue!10}{\textbf{95.2}} & \cellcolor{blue!10}{\textbf{98.0}} & \cellcolor{blue!10}{\textbf{55.6}} & \cellcolor{blue!10}{\textbf{81.2}} & \cellcolor{blue!10}{\textbf{89.5}} & \cellcolor{blue!10}{\textbf{63.1}} & \cellcolor{blue!10}{\textbf{87.2}} & \cellcolor{blue!10}{\textbf{92.3}} & \cellcolor{blue!10}{\textbf{38.2}} & \cellcolor{blue!10}{\textbf{67.4}} & \cellcolor{blue!10}{\textbf{78.5}} \\
        \bottomrule
    \end{tabularx}
    \vspace{1pt}
    \caption{Zero-shot \& Fine-tuned image--text retrieval on the test splits of Flickr30k and COCO Captions with models pre-trained on YFCC15M. ViT-B/32 is used for all setup.}
    \label{result:yfcc15m_retrieval}
\end{table*}
For multi-modal evaluations, we test both the zero-shot and fine-tuned image--text (and text--image) retrieval on Flickr30k and COCO Captions benchmarks.
Image-text pairs are ranked according to their similarity scores.
Tab.~\ref{result:yfcc15m_retrieval} shows the performance for image--text retrieval tasks of MCD pretrained on YFCC15M dataset.
Our MCD outperforms all state-of-the-art baselines across every measure with a considerable margin.
By incorporating a log-ratio loss with metric learning characteristics into the CLIP framework, our proposed approach has achieved significant improvements in image-text retrieval performance.

\paragraph{Vision--Language Compositionality.}
\begin{table}[t]
    \tabcolsep=0.18cm
    \small
    \centering
    \begin{tabularx}{\linewidth}{l |cc| ccc}
        \toprule
        Method & Aug. & Misalign & Replace & Swap & Add  \\
        \midrule
        CLIP~\cite{radford2021clip} & & N/A & 73.6	& 59.5 & 69.4 \\
        SLIP~\cite{mu2022slip} & \checkmark & N/A & 74.7	& 58.6 & 69.1 \\
        DeCLIP~\cite{li2022declip} & \checkmark & Disregard & 74.5 & 58.2 & 66.8 \\
        UniCLIP~\cite{lee2022uniclip} & \checkmark & $f_A$ & 75.5 & 58.4 & 70.4 \\
        MCD (Ours) & \checkmark & $\mathcal{L}_{\text{D}}$ & \cellcolor{blue!10}{\textbf{76.2}} & \cellcolor{blue!10}{\textbf{61.5}} & \cellcolor{blue!10}{\textbf{71.7}} \\
        \bottomrule
    \end{tabularx}
    \vspace{1pt}
    \caption{Evaluation on SugarCrepe~\cite{hsieh2023sugarcrepe}. All models were pre-trained on YFCC15M with ViT-B/32 backbone. Models except CLIP involve random image augmentations (Aug.) with different schemes for dealing with image--text misalignments (Misalign).
    While previous methods either disregard the issue~\cite{li2022declip} or introduce an additional augmentation encoder ($f_A$)~\cite{lee2022uniclip}, MCD manages to harness the misalignment for the distillation loss ($\mathcal{L}_{\text{D}}$).
    }
    \label{result:sugarcrepe}
\end{table}

To conduct a thorough analysis of multimodal representation learning, we assess the performance of our model using the SugarCrepe~\cite{hsieh2023sugarcrepe} dataset.
This dataset serves as a de-biased benchmark specifically designed for evaluating the compositionality aspect of vision-language models.
SugarCrepe introduces a set of challenging negative captions for COCO image-text pairs by replacing, swapping, or adding certain concepts to the ground truth caption, and gauge the model's capability to discern the positive from its distractor.
Tab.~\ref{result:sugarcrepe} summarizes the result, showcasing the performance of various models pretrained on the YFCC15M dataset. 
Our findings demonstrate that the MCD exhibits significantly better compositionality compared to prior methods.
This improved performance is attributed to leveraging augmentations with proper handling for misalignment arising from these augmentations.
While other methods often show improved performance over CLIP baseline, adopting augmentations without meticulous management of misalignment cannot maximize their utility.

\subsection{MCD Pretraining on CC3M Dataset}
\label{sec:cc3m_comparison}
In this section, we compare MCD against other state-of-the-art Contrastive Language-Image Pretraining approaches~\cite{radford2021clip, mu2022slip,li2022declip}.
All models are pretrained on the CC3M dataset with a learning rate 5e-4 for 40 epochs\footnote{More detailed training configuration will be provided in supplement.}.
Tab.~\ref{tab:cc3m} shows the ImageNet zero-shot results of MCD with other CLIP variants.
MCD outperforms all CLIP variants without external training sources such as Nearest Neighbor supervision with large memory queues (NNS) or augmentation information during training (AUG).
Furthermore, MCD does not require any additional parameters for the SSL projection layer~\cite{mu2022slip,li2022declip} or additional network for augmentation-aware feature embedding~\cite{lee2022uniclip}.
\begin{table}[t!]
  \centering
  \small
  \begin{tabularx}{\columnwidth}{l |c| c c c}
    \toprule
    Method & Encoder & SSL & EXT & Top1(\%) \\ \hline \\[-9pt]
    CLIP~\cite{radford2021clip} & \multirow{5}{*}{ViT-B/16} & - & - & 19.6 \\
    SLIP~\cite{mu2022slip} & & SimCLR~\cite{chen2020simclr} & - & 23.2 \\
    DeCLIP~\cite{li2022declip} & & SimSiam~\cite{chen2021simsiam} & NNS & 25.4 \\
    UniCLIP~\cite{lee2022uniclip} & & MP-NCE~\cite{lee2022uniclip} & AUG & 27.8 \\
    MCD (Ours) & & MP-NCE~\cite{lee2022uniclip} & - & \textbf{28.2} \\ 
    \bottomrule
  \end{tabularx}
  \vspace{1pt}
  \caption{
  ImageNet-1k Top 1 zero shot accuracy with models pretrained on CC3M dataset. SSL denotes the vision self-supervision term used in each model and EXT denotes external sources involved during training.
  NNS is nearest neighbor supervision using a separate memory queue, and AUG is the vectorized information of each random augmentation conducted during training.
  }
  \label{tab:cc3m}
\end{table}
\subsection{Ablation Study}

This section presents ablation studies to evaluate the contribution of each component in our proposed approach, MCD, towards the final performance. To this end, we pre-train all models on the YFCC15M dataset and evaluate them using zero-shot learning on the Imagenet-1k validation set. Specifically, we implement the MP-NCE loss without augmentation encoding, which results in an accuracy of 39.6. Our results in Tab.~\ref{tab:ablation} demonstrate that each loss component in MCD has a positive impact on the final performance, leading to an overall improvement in accuracy. 
These findings highlight the importance of each component in our proposed approach and validate its effectiveness in improving the zero-shot classification performance.
Note that MCD outperforms (b) (\ie, UniCLIP) that explicitly includes the augmentation information during training with only $\mathcal{L}^\text{pos}$, showing the effectiveness of harnessing the misalignments that occur during random image augmentation for training.

\begin{table}[t]
\tabcolsep=0.3cm
  \centering
  \small
\resizebox{1.0\linewidth}{!}{
  \begin{tabular}{c c c c c c c}
    \toprule
    & $\mathcal{L}^{\text{base}}$ & $f_A$ & $\mathcal{L}^{\text{pos}}$  & $\mathcal{L}^{\text{neg}}$ & $\mathcal{L}^{\text{noisy}}$ & Top1 Acc (\%) \\ \hline
    (a) & \checkmark & & & & & 39.6 \\
    (b) &\checkmark & \checkmark & & & & 42.8 \\
    (c) & \checkmark & & \checkmark & & & \textbf{43.9} \\
    (d) & \checkmark & & \checkmark & \checkmark & & \textbf{44.3} \\
    (e) & \checkmark & & \checkmark & \checkmark & \checkmark & \textbf{44.7} \\
    \bottomrule
  \end{tabular}}
  \vspace{1pt}
  \caption{
  Ablation study on ImageNet-1k Top 1 zero shot accuracy for vision-language pretraining for each loss components of MCD. All models were pretrained with a ViT-B/32 backbone with a basic contrastive loss $\mathcal{L}^C$ in Eq~(\ref{eq:MP-NCE}) and MLM loss in Eq~(\ref{eq:MLM}), which we abbreviate as $\mathcal{L}^{\text{base}}$.
  All $\mathcal{L}^{\text{pos}}$, $\mathcal{L}^{\text{neg}}$, $\mathcal{L}^{\text{noisy}}$ shows a consistent gain in zero-shot performance.
  $f_A$ denotes the augmentation encoder, making (b) analogous to UniCLIP~\cite{lee2022uniclip}.
  }
  \label{tab:ablation}
  \vspace{-5pt}
\end{table}
\section{Conclusion}

We propose MCD, a new training strategy for dealing with misalignments occured by random image augmentations under visual--language pretraining.
Our novel distillation formulation enables data-efficient training under Contrastive Language-Image Pretraining.
Future works will include extending MCD frameworks to other modalities.
\newpage

{\small
\bibliographystyle{ieee_fullname}
\bibliography{iccv2024}
}

\end{document}


\title{Appendix for Misalign, Contrast, then Distill:\\Rethinking Misalignments in Language-Image Pretraining}

\author{First Author\\
Institution1\\
Institution1 address\\
{\tt\small firstauthor@i1.org}
\and
Second Author\\
Institution2\\
First line of institution2 address\\
{\tt\small secondauthor@i2.org}
}

\maketitle
\ificcvfinal\thispagestyle{empty}\fi

\section{Implementation Details}
In this section, we describe the implementation details of each experimental setup in our main paper accordingly.
For CC3M, All models are trained with a ViT-B/16 backbone with a learning rate of 5e-4 and weight decay of 0.5.
For YFCC15M, we train the model with ViT-B/32 backbone, batch size 4096, learning rate 1e-3, and weight decay 0.2.

\paragraph{Compatibility with Other Supervisions.}
In our main paper, we compare the ImageNet zero-shot Top1 accuracy between MCD and other CLIP variants, where MCD is only given the additional supervision of augmented images.

\begin{table}[h!]
  \centering
  \small
  \begin{tabularx}{\columnwidth}{l c c c c c}
    \toprule
    Method & VLC & MLM & EDA & NNB & Top1(\%) \\ \hline \\[-9pt]
    DeCLIP~\cite{li2022declip} & \checkmark & \checkmark & \checkmark & \checkmark & 41.2 \\
    MCD & \checkmark & & & & \textbf{43.4} \\
    MCD & \checkmark & \checkmark & & & 44.3 \\
    MCD & \checkmark & \checkmark & \checkmark & & 44.5 \\
    MCD & \checkmark & \checkmark & \checkmark & \checkmark & \textbf{44.9} \\
    \bottomrule
  \end{tabularx}
  \caption{
  ImageNet-1k Top 1 zero shot accuracy with models pretrained on YFCC15M dataset under additional supervision during training (\eg, VLC: Vision-Language Contrastive Loss, MLM: Masked Language Modeling, EDA: Text augmentation with EDA~\cite{wei2019eda}, NNB: Nearest Neighbor Memory Bank~\cite{li2022declip}).
  It can be seen that previous additional supervision are consistently compatible with our MCD.
  }
  \label{tab:cc3m}
\end{table}

\paragraph{Experimental Setups for Main Paper Table 4.}
In Tab. 4 of our main paper, we compare the ImageNet zero-shot Top1 accuracy between MCD, other CLIP variants and its KL-distillation version which are pre-trained on the CC3M dataset.
\begin{itemize}[leftmargin=10pt]
    \item (1) CLIP : The reported CLIP performance in SLIP~\cite{mu2022slip}\footnote{https://github.com/facebookresearch/SLIP}. Only the original image-text pairs are given without any additional augmented views for either image or text. In this setting, the final loss is the InfoNCE loss with a batch size 2048. We here also report the performance when trained with an identical batch size with others (\ie, 1024), which will be 18.32.
    \item (2) SLIP : In this setting, SimCLR~\cite{chen2020simclr} loss of two random augmented image views is additionally used along with CLIP loss. For SimCLR, three Multi-Layer Perceptrons (with weights $\in\mathbb{R}^{D\times 4096}, \mathbb{R}^{4096\times 4096}, \mathbb{R}^{4096\times 256}$)\footnote{$D$: output dimension of the ViT. For ViT-B, $D=768$.} are applied along with Batch Norm and ReLU operation. In this setting, models are trained with batch size 1024.
    \item (3) KL w/o SSL : Compared to our MCD, the image-text misalignments can also be distilled by a simple KL divergence loss. Following previous works~\cite{caron2021dino}, we use a temperature 0.04 for the teacher, 0.1 for the student, momentum centering 0.9, and momentum weight gradually increasing from 0.994 to 1 in a cosine schedule. Models are trained with a batch size 1024.
    \item (4) KL w/ SSL : To match the level of supervision with MCD, we add self-supervision terms (\ie, MP-NCE) on top of the language--image contrastive loss in (1) and distillation loss of (3). Models are trained with a batch size 1024. Note that despite MCD (28.02) outperforms all the distillation variants, validating the effectiveness of our proposed method.
\end{itemize}


\paragraph{Linear Probing.}
To implement linear probe evaluation, we follow CLIP~\cite{radford2021clip} to train a logistic regression classifier on the frozen visual features extracted by the image encoder.
Specifically, we train the logistic regression classifier using L-BFGS algorithm provided by scikit-learn with maximum 1,000 iterations, and report the corresponding metric for each dataset\footnote{https://github.com/facebookresearch/SLIP/blob/main/main\_linear.py}.
Parameters for L2 regularization are determined using hyperparameter sweep on the validation sets.
Standard cropping and flipping augmentations~\cite{szegedy2015going} are used for linear probing.


\section{Dataset Specification}

\paragraph{Downstream datasets.}
We evaluate the transferability of MCD on 11 downstream classification datasets including ImageNet~\cite{Russakovsky2014imagenet1k}, Oxford-IIIT Pets~\cite{parkhi2012cats}, CIFAR-10~\cite{krizhevsky2009learning}, CIFAR-100~\cite{krizhevsky2009learning}, Describable
Textures~\cite{cimpoi2014describing}, Stanford Cars~\cite{krause20133d}, Food-101~\cite{bossard2014food}, Oxford Flowers 102~\cite{nilsback2008automated}, FGVC Aircraft~\cite{maji2013fine},
SUN397~\cite{xiao2016sun} and Caltech-101~\cite{fei2004learning}.
The details of each dataset are listed in Tab.~\ref{tab:downstream} and we follow
the same data split and evaluation metric as CLIP for a fair comparison.
\begin{table}[t]
  \centering
  \footnotesize
  \begin{tabular}{l c c c c}
    \toprule
    Dataset & \# class & \# train & \# test & Eval Metric \\ \midrule
    CIFAR-10 & 10 & 50,000 & 10,000 & accuracy \\
    CIFAR-100 & 100 & 50,000 & 10,000 & accuracy \\
    Describable Textures & 47 & 3,760 & 1,880 & accuracy \\
    Stanford Cars & 196 & 8,144 & 8,041 & accuracy \\
    Food-101 & 101 & 75,750 & 25,250 & accuracy \\
    Oxford-IIIT Pets & 37 & 3,680 & 3,669 & mean per class \\
    Oxford Flowers 102 & 102 & 2,040 & 6,149 & mean per class \\
    FGVC Aircraft & 100 & 6,667 & 3,333 & mean per class \\
    SUN397 & 397 & 19,850 & 19,850 & accuracy \\
    Caltech-101 & 102 & 3,060 & 6,085 & mean per class \\
    ImageNet & 1000 & 1,281,167 & 50,000 & accuracy \\
    \bottomrule
  \end{tabular}
  \caption{
  ImageNet-1k Top 1 accuracy for models pretrained on the YFCC15M dataset with ViT-B/32 backbone .
  MCD shows a consistent tendency across various scales of datasets.
  }
  \label{tab:downstream}
\end{table}

{\small
\bibliographystyle{ieee_fullname}
\bibliography{egbib}
}